\documentclass[runningheads]{llncs}

\usepackage{eccv}

\usepackage{eccvabbrv}

\usepackage{graphicx}
\usepackage{booktabs}

\usepackage[accsupp]{axessibility}  % Improves PDF readability for those with disabilities.

\usepackage{hyperref}

\usepackage{orcidlink}

\begin{document}

\title{White Aggregation and Restoration for \\Few-shot 3D Point Cloud Semantic Segmentation} 

\titlerunning{White Aggregation and Restoration for FS-PCS}


% Second Author\inst{2,3}\orcidlink{1111-2222-3333-4444} \and
% Third Author\inst{3}\orcidlink{2222--3333-4444-5555}}

\author{Jiyun Im \and
SuBeen Lee \and
Miso Lee \and
Jae-Pil Heo*}

\def\thefootnote{*}\footnotetext{Corresponding author}
% \footnotetext{\href{https://github.com/JiyunIm00/WARM.git}{Codes.}}

\authorrunning{J. Im et al.}

\institute{Sungkyunkwan University \\
\email{\{bbangsil0110, leesb7426, dlalth557, jaepilheo\}@skku.edu}}

\maketitle

\begin{abstract}
\label{sec:abstract}
Few-shot 3D Point Cloud Semantic Segmentation (FS-PCS) aims to predict per-point labels for an unlabeled point cloud, given only a few labeled examples.
To extract representations from the limited labeled set, existing methods have constructed prototypes with Farthest Point Sampling (FPS).
However, we found that this convention results in performance instability due to its sensitivity to FPS-induced variations, while the prototype generation process remains underexplored in the field.
This motivates us to investigate deterministic prototype generation method based on attention mechanism.
Despite its potential, we found that vanilla attention module suffers from the distributional gap between prototypical tokens and support features.
To overcome this, we provide a simple approach,  White Aggregation and Restoration Module (WARM), which resolves the misalignment by wrapping cross-attention with whitening and coloring transformations.
Specifically, whitening alig-ns the features to tokens before the attention process, and coloring subsequently restores the original distribution to the attended tokens.
This design enables robust attention, thereby generating prototypes that capture the semantic relationships in support features.
% WARM achieves state-of-the-art performance with a significant margin on FS-PCS benchmarks, and demonstrates its effectiveness through extensive experiments.
WARM achieves state-of-the-art performance with a significant margin on the S3DIS dataset, and competitive performance on the ScanNet dataset.
Further experiments demonstrate its effectiveness in deterministic prototype generation.
Code is publicly available at: \href{https://github.com/JiyunIm00/WARM.git}{https://github.com/JiyunIm00/WARM.git}

\end{abstract}

\section{Introduction}
\label{sec:intro}

Understanding the semantics of 3D point clouds has become crucial as its applications have advanced \cite{xiao2023unsupervised, betsas2025deep}.
Although recent methods have achieved remarkable performance, they rely on large amounts of labeled data, which requires expensive labor \cite{armeni20163d, dai2017scannet}.
To alleviate this data reliance, Few-shot 3D Point Cloud Semantic Segmentation (FS-PCS) was introduced \cite{zhao2021few}.
It aims to segment unseen novel classes in unlabeled point clouds, referred to as the query set, using a small number of labeled examples, referred to as the support set.

To fully utilize the information from the support set, previous studies have constructed class prototypes \cite{zhao2021few, an2024multimodality, an2024rethinking, ning2023boosting, PAPFZS3D}.
It is widely adopted in other few-shot downstream tasks, where numerous studies focus on improving representation quality due to its impact on performance \cite{liu2020part, zhang2022feature, fan2022self, lee2025temporal}.
In FS-PCS, most existing methods have relied on algorithmic approach such as Farthest Point Sampling (FPS) to construct multiple prototypes of the target class, following \cite{zhao2021few}.

\begin{figure}[!t]
    \begin{minipage}[T]{0.42\columnwidth}
        \centering
        \renewcommand{\arraystretch}{1.1} % Default value: 1
\setlength{\tabcolsep}{1.5pt}
\captionof{table}{
mIoU~(\%) distribution resulting from random seeds.
Performance is evaluated under the 1-way 1-shot setting on the first split of S3DIS~\cite{armeni20163d}.
`FPS + \textit{min-dist.}' denotes a simple baseline that assigns labels based on the minimum distance to FPS-constructed class prototypes.
The distribution of mIoUs show the sensitivity of FS-PCS models to FPS results, that even the complex segmentation structure~\cite{an2024rethinking} falls short to generalize.
On the other hand, attention-based WARM shows much stability.
}
\medskip
\label{tabs:1_motivation}
\centering
\scriptsize
\begin{tabular}{l|c|c|c|c}
\toprule
{Method} & Max & Min & Mean & Std. \\
\midrule
COSeg~\cite{an2024rethinking} & 52.86 & 37.99 & 45.67 & 2.41 \\ 
FPS + \textit{min-dist.} & 52.14 & 46.86 & 49.48 & 0.86 \\
\midrule
WARM & 60.16 & 60.16 & 60.16 & - \\ 
\bottomrule
\end{tabular}
    \end{minipage}
    \hfill
    \begin{minipage}[T]{0.55\columnwidth}
        \includegraphics[width=\columnwidth]{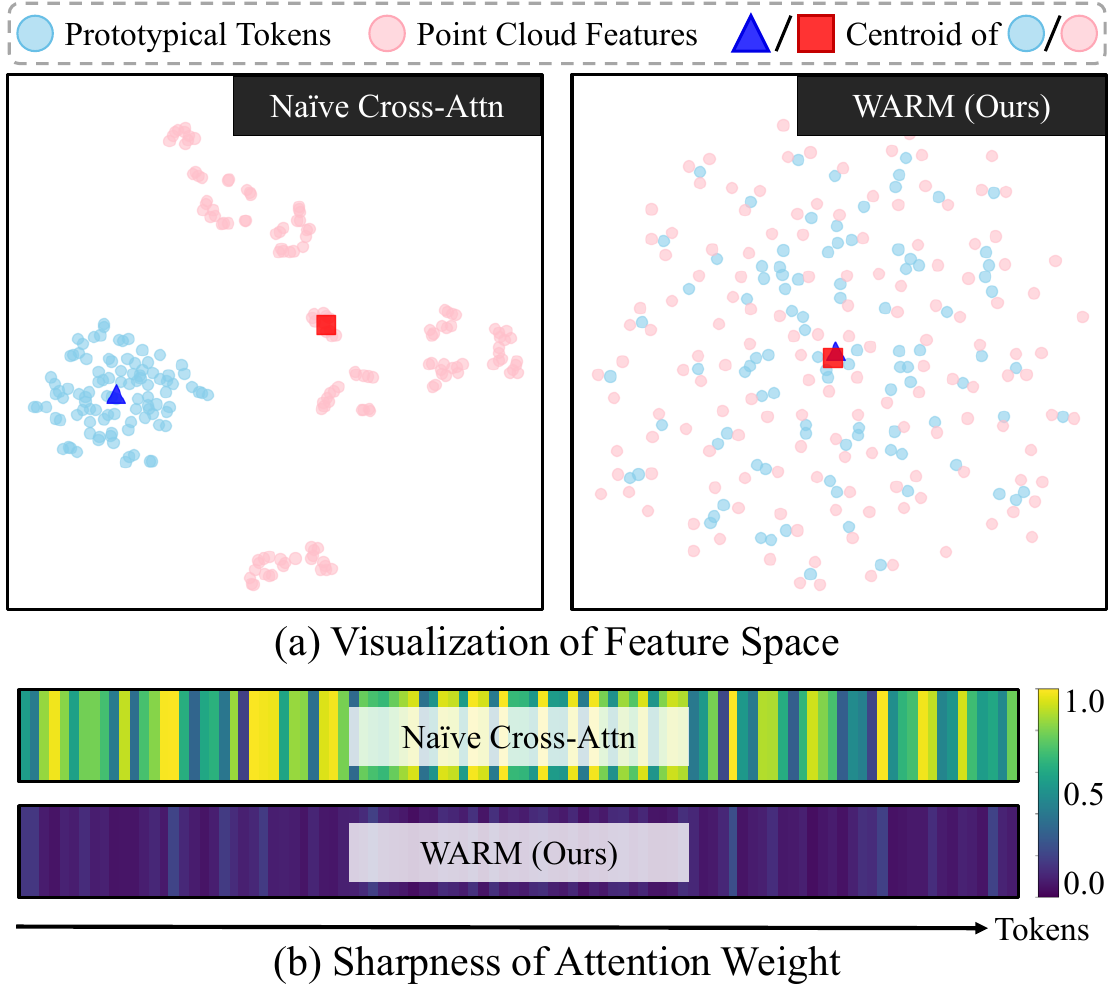}
        \caption{
        Comparison of na\"ive cross-attention and WARM in prototype construction.
        (a) t-SNE visualization of the alignment between prototypical tokens (queries) and point cloud features (keys).
        (b) Attention sharpness measured via attention entropy~\cite{zhai2023attentionentropy}, averaged for each token. 
        }
        \label{figs:1_motivation}
    \end{minipage}
\end{figure}

Surprisingly, these conventional prototypes cause performance instability as shown in \cref{tabs:1_motivation}.
Due to its results being dependent on the initially selected point \cite{yang2019modeling}, FPS produces inconsistent results that create fluctuations in performance.
Particularly, the subsequent sophisticated module \cite{an2024rethinking} aggravates this sensitivity when compared to a simple distance-based method, highlighting the importance of prototype construction in FS-PCS.
This motivates us to devise an advanced, consistent prototype generation module.

One may consider the attention mechanism \cite{vaswani2017attention} as a promising candidate, given its deterministic nature and demonstrated adaptability to prototype generation \cite{zhang2022feature, cheng2022masked, carion2020end, lee2025temporal}.
Nevertheless, we find that attention alone remains insufficient in the case of FS-PCS.
Ideally, the cross-attention module should be optimized such that the prototypical tokens adaptively aggregate information from support features to form representative prototypes.
However, this process fails due to a large distributional gap between the prototypical tokens and the support point features, as shown in \cref{figs:1_motivation} (a).

This misalignment prevents the prototypical tokens from establishing meaningful interactions across semantically correlated regions, causing attention to collapse onto a limited subset of features and consequently degrading representation quality as discussed in \cite{kovaleva2019darkbert, dong2021attentionisnot, zhai2023attentionentropy, lee2024activating}.
Consistently, in \cref{figs:1_motivation} (b), most tokens allocate more than half of their total attention weight to an individual support point, indicating a lack of compositional understanding of the class.

To enable effective attention-based prototype generation in FS-PCS, we introduce a simple approach that integrates cross-attention with established alignment techniques.
Our proposed White Aggregation and Restoration Module (WARM) constructs semantically meaningful prototypes via robust attention induced by aligned prototypical tokens and support features.
Specifically, in order to align with the prototypical tokens, the support features are transformed to a whitened space \cite{bell1997independent} where it is standardized and decorrelated.
By operating in a shared feature space, the tokens establish interactions with the features that promote semantically coherent representations instead of ones locally confined in the projection space.
Moreover, decorrelating features promotes smoother and more stable optimization \cite{daneshmand2020batch, huang2018decorrelated}.
Subsequently, coloring—the inverse operation of whitening—restores the separated characteristics to the resulting prototypes, preserving essential feature properties for faithful representation.
Ultimately, WARM contributes to effective attention-based prototype generation in FS-PCS by improving compatibility between prototypical tokens and support features, while also fostering stable optimization.

As a result, our attention-based WARM alone achieves competitive performance on FS-PCS benchmarks \cite{armeni20163d, dai2017scannet}, with state-of-the-art performance on S3DIS dataset.
We validate our approach through extensive experiments, demonstrating its effectiveness in addressing the challenges of adapting attention-based prototype generation to FS-PCS.
The contribution of our work lies in rethinking a simple architecture in light of task-specific challenge and introducing a complementary perspective for future research in FS-PCS.
In summary, our contributions are as follows:
\begin{itemize}
    \item We identify the challenges inherent in adapting attention mechanisms for prototype generation in FS-PCS, providing detailed analysis of the distributional factors that limit their representational capacity.
    \item We devise a simple solution, WARM, that utilizes concrete alignment method to address the distributional disparity during cross-attention, and validate its effectiveness through comprehensive ablation analyses.
    \item Our method achieves competitive performance in FS-PCS benchmarks, without the aid of complex decoders.
\end{itemize}

\section{Related Works}
\label{sec:related}

\subsection{Few-Shot 3D Point Cloud Segmentation}
Few-Shot 3D Point Cloud Segmentation (FS-PCS) aims to predict per-point labels for a query point cloud given a small set of labeled support samples.
Following the prototypical paradigm \cite{snell2017prototypical}, prior FS-PCS methods construct prototypes to represent the support set information.
Given the limited supervision, the expressiveness of these prototypes is crucial to performance.
Most existing works adopt the multi-prototype pipeline introduced by \cite{zhao2021few}, where seeds are selected via Farthest Point Sampling (FPS) in coordinate \cite{an2024rethinking, an2024multimodality} or feature space \cite{zhao2021few}, followed by clustering the surrounding point cloud or multi-modal \cite{an2024multimodality} features to form prototypes.
Alternatively, some methods use a single prototype derived from masked average pooling \cite{PAPFZS3D}.

While these approaches are intuitive, they rely on hand-crafted rules, and most studies \cite{ning2023boosting, li2024localization} focus on adapting such prototypes to the query rather than improving the prototype construction itself.
In contrast, we explore a learnable alternative based on attention mechanisms to enhance the flexibility and expressiveness of prototype generation.

\subsection{Attention-based Prototype Generation}
In the image domain, it is common to construct prototypes using learnable prototypical tokens through cross-attention mechanisms, such as DETR \cite{carion2020end} and Mask2Former \cite{cheng2022masked}.
In contrast, 3D instance segmentation and object detection typically adopt non-learnable, non-parametric tokens sampled directly from the input point cloud to summarize features via attention \cite{liu2022masked, misra2021end, schult2022mask3d}.
Although some work in 3D domain leverage parametric tokens \cite{wang2022detr3d, xie2023pixel}, they rely on auxiliary information such as camera coordinates or 2D image features.
This tendency primarily arises from the inherent difficulty of aligning the point cloud distribution with parameter initializations \cite{zhou2024dynamic}.
Through our work, we inspect the underlying matter that induces the misalignment between point cloud features and prototypical tokens, and propose a simple approach that enables attention-based prototype generation in FS-PCS.

\section{Preliminary}
\label{sec:preliminary}

% \subsection{Preprocessing Protocols}
% \label{subsec:protocols}
% FS-PCS studies are conducted under two distinct strategies of preprocessing raw point clouds, 1) \textit{foreground oversampling} \cite{zhao2021few, wang2025taylor, wang2025dypolyseg} and 2) \textit{uniform sampling} \cite{an2024rethinking, an2024multimodality}.
% As noted in \cite{an2024rethinking}, the former relies on ground-truth masks to over-sample foreground regions, limiting its use in real-world settings where annotations are unavailable.
% Therefore, we adopt uniform sampling, following \cite{an2024rethinking}.

\subsection{Problem Formulation}
FS-PCS aims to segment unlabeled point clouds based on a small set of labeled points.
Specifically, in each $N$-way $K$-shot episode following the well-known meta-learning paradigm \cite{vinyals2016matching}, it consists of $K$ labeled point clouds for $N$ classes, called the support set $S=\{ \{ (X^{S}_{n,k}, Y^{S}_{n,k}) \}^K_{k=1} \}^N_{n=1}$ and $U$ unlabeled ones, named the query set $Q=\{(X^Q_u, Y^Q_u)\}^U_{u=1}$.
Here, $X$ and $Y$ represent the point cloud and its corresponding mask, respectively.
As a result, the goal of FS-PCS is to predict per-point semantic labels for each query point cloud in $Q$, leveraging the information in the support set $S$.
Note that training and testing utilize mutually exclusive class sets, $\mathcal{C}_\text{base}$ and $\mathcal{C}_\text{novel}$, individually, \textit{i.e.}, $\mathcal{C}_\text{base} \cap \mathcal{C}_\text{novel} = \varnothing$.
In the below sections, we assume a 1-way 1-shot setting for better clarity.

\subsection{Farthest Point Sampling (FPS)}
\label{subsec:fps}
Given a point cloud $X \in \mathbb{R}^{L \times 3}$, we extract the point-wise features $F \in \mathbb{R}^{L \times D}$, where $L$ and $D$ denote the number of points and the feature dimension, respectively.
For simplicity, we assume that both the support and query sets contain $L$ points.
We divide the support set into two semantic classes, foreground (FG) and background (BG), based on the ground-truth mask $Y^S$.
Therefore, we obtain the class-specific features $F^C \in \mathbb{R}^{L^C \times D}$ for each class $C \in \{\text{FG}, \text{BG}\}$ within the support features $F^S$, where $L^C$ denotes the number of points of the class $C$.

To effectively leverage the information within a few labeled samples, FS-PCS methods typically construct prototypes using the Farthest Point Sampling (FPS) algorithm \cite{an2024multimodality, an2024rethinking, li2024localization, zhao2021few}.
The FPS algorithm starts from a randomly selected $l$-th point feature $F^C_{l}$ for each class, which serves as the initial element $r^C_1$ of the subset.
Then, it iteratively expands the subset $R^C_{t} = \{r^C_{1}, \cdots, r^C_{t}\}$ at each step $t = 1, \cdots, T$, as follows:
\begin{equation}
    \begin{split}
        \label{eq:fps}
        r^C_{t+1} = \underset{f^C \in F^C \setminus R^C_{t}}{\arg\max} \big( \underset{r^C \in R^C_{t}}{\min} \left\| f^C - r^C \right\|_2 \big).
    \end{split}
\end{equation}
The resulting subset $R^C_{T}$ serves as a compact set of representative point features for each class.

\section{Motivation}
\label{sec:motivation}

\subsection{Attention-based Prototype Generation}
\label{subsec:attention_based_prototype_generation}
Despite the wide adoption of FPS-based prototypes in FS-PCS, it suffers from intrinsic randomness.
As detailed in \cref{subsec:fps}, it initiates by randomly selecting a point from the input, then iteratively chooses the farthest point from the previously selected point set.
Therefore the resulting set is highly dependent on the random choice, leading to instability in performance, as shown in \cref{tabs:1_motivation}.

In contrast to algorithmic approaches in FS-PCS, other few-shot downstream tasks \cite{zhang2022feature, liu2020part, lee2025temporal} have developed prototype generation techniques that leverage the cross-attention mechanism.
These attention-based strategies offer notable advantages over the FPS-based ones.
First, they generate deterministic outputs, eliminating stochastic effects and ensuring consistent performance.
Secondly, the attention mechanisms are fully parallelizable via matrix multiplications, unlike the inherently sequential nature of FPS.
Furthermore, they are task-optimizable, as they constitute learnable components rather than fixed sampling heuristics like FPS.
These strengths motivate us to adopt attention-based prototype generation in FS-PCS.

Given $M$ prototypical tokens $P=[P_1,P_2, \cdots, P_M]$ where each token $P_m \in \mathbb{R}^D$, we can construct support prototypes $\hat{P}^C \in \mathbb{R}^{M \times D}$ for each class, as follows:

\begin{equation}
    \hat{P}^C_{m} = P_m +
    \sum_{l \in L^C} 
    \frac{
    \exp \left( W_q(P_m)W_k(F^C_l) \right)
    }{
    \sum_{l^\prime \in L^C} 
    \exp \left( W_q(P_m)W_k(F^C_{l^\prime}) \right)
    } W_v(F^C_l),
    \label{eq:vanilla_ca}
\end{equation}
where $W_q(\cdot)$, $W_k(\cdot)$, and $W_v(\cdot)$ are projection layers for the \textit{query}, \textit{key}, and \textit{value}, respectively.

\subsection{Attention Misalignment in FS-PCS}
\label{subsec:attention_misalignment}
To construct representative prototypes using the cross-attention mechanism, it is crucial that the attention in \cref{eq:vanilla_ca} effectively captures the underlying semantic relationships among the support features.
Otherwise, the whole process is prone to yield sub-optimal outputs, often relying heavily on skip connections \cite{dong2021attentionisnot,zhai2023attentionentropy,lee2024activating}.
For this to hold, the support features and prototypical tokens must reside in a well-aligned embedding space where semantic similarity is reflected by feature proximity.

In order to assess the robustness of attention-mechanism in FS-PCS, we analyze the vanilla cross-attention module in terms of the distribution in the embedding space of $P$ and $F^\text{FG}$.
Note that our analysis is derived by averaging across test episodes from the 1-way 1-shot setting on the first split of the S3DIS \cite{armeni20163d} dataset with a focus on the foreground (FG) class, as the background (BG) often contains multiple semantic categories that could introduce confounding factors.

\begingroup
\setlength{\tabcolsep}{4.0pt}
\renewcommand{\arraystretch}{1.1}
\begin{table}[t]
\caption{Distributional discrepancy analysis. $P$ and $F^\text{FG}$ denote prototypical tokens and support foreground (FG) features, respectively, while $W_q(P)$ and $W_k(F^\text{FG})$ are projected versions of them using the attention layer.
(a) $\text{Dist}(\cdot,\cdot)$ measures the degree of misalignment between two matrices by computing their average distance, whereas (b) corresponds to the distance between their means.
(c) $\mathcal{D}^\text{instance}$ represents the feature scale within each instance, while (d) $\kappa^\text{instance}$ quantifies anisotropy.
}
\footnotesize
\centering
\begin{tabular}{l|cc|cc}
    \toprule
    & $P$ & $F^\text{FG}$ & $W_q(P)$ & $W_k(F^\text{FG})$
    \\
    \midrule
    (a)~Distance of two matrices~$\text{Dist}(\cdot, \cdot)$~(\cref{eq:q_k_distance}) & \multicolumn{2}{c|}{282.61} & \multicolumn{2}{c}{271.18}
    \\
    \midrule
    (b)~Distance of mean~$\text{Dist}_\mu(\cdot, \cdot)$~(\cref{eq:mu_distance}) & \multicolumn{2}{c|}{264.17} & \multicolumn{2}{c}{251.56}
    \\
    \midrule
    (c)~Dipersion~$\mathcal{D}^\text{instance}$~(\cref{eq:dispersion}) & 1.14 & 94.68 & 1.12 & 95.40
    \\
    \midrule
    (d)~Anisotropy~$\kappa^\text{instance}$~(\cref{eq:anisotropy}) & 5.36 & 1122.81 & 10.41 & 12392.23
    \\
    \bottomrule
\end{tabular}
\label{tabs:2_misalignment}
\end{table}
\endgroup

First, the simplest metric to quantify the discrepancy between the distributions of $F^\text{FG}$ and $P$ is the average distance, denoted as:
\begin{equation}
    \text{Dist}(P,F^\text{FG}) = 
    \frac{1}{L^\text{FG}}
    \sum_{l=1}^{L^\text{FG}} 
    \min_m \parallel P_m - F^\text{FG}_{l} \parallel_2.
    \label{eq:q_k_distance}
\end{equation}
To better understand the contributing factors in $\text{Dist}(\cdot , \cdot)$ presented in \cref{tabs:2_misalignment}~(a), we introduce a related metric that measures the distance between the mean vectors:
\begin{equation}
    \text{Dist}_{\mu}(P,F^\text{FG}) =  
    \parallel \mu^P - \mu^\text{FG} \parallel_2,
    \label{eq:mu_distance}
\end{equation}
where $\mu^{\text{FG}}=\frac{1}{L^{\text{FG}}} \sum_{l=1}^{L^{\text{FG}}} F^{\text{FG}}_{l} \in \mathbb{R}^{1 \times D}$ is the mean of support FG features, and similarly $\mu^P$ the mean of $P$.
\cref{tabs:2_misalignment} (a) and (b) imply that a substantial portion of $\text{Dist}(\cdot , \cdot)$ arises from the discrepancy between the means of $F^\text{FG}$ and $P$.
Therefore, by aligning the centers, the misalignment is expected to be largely mitigated.

Nevertheless, even with the centers aligned, additional components of misalignment still exist.
For instance, if the features $F^\text{FG}$ exhibit substantially larger dispersion around the center than the tokens $P$, this scale discrepancy can hinder effective semantic aggregation.
Conversely, pronounced anisotropy in the feature directions may still restrict the tokens from fully capturing the semantic context, despite comparable scale.
To inspect these factors, we first define the within-instance dispersion $\mathcal{D}^{\text{instance}}$, to describe the scale of the feature space:
\begin{equation}
        \mathcal{D}^\text{instance} = 
        \frac{1}{L^\text{FG}} 
        \sum_{l=1}^{L^\text{FG}} 
        \parallel 
        F^\text{FG}_{l} - \mu^{\text{FG}} 
        \parallel_2.
        \label{eq:dispersion}
\end{equation}
Next, to compare the geometric differences, we measure instance-wise anisotropy $\kappa^{\text{instance}}$, defined as follows:
\begin{equation}
    \label{eq:anisotropy}
    \begin{gathered}
        \kappa^{\text{instance}} = \Sigma^\text{FG}_{11} / \min \{ \Sigma^\text{FG}_{ii} \mid 1 \leq i \leq r, 0 < \Sigma^\text{FG}_{ii} \}
        ,
        \\
        \text{diag}\left(\Sigma^\text{FG}\right) = 
        \left[
        \Sigma^\text{FG}_{11}, 
        \Sigma^\text{FG}_{22}, 
        \cdots,
        \Sigma^\text{FG}_{rr}
        \right]
        ,\quad
        r =\min\left(L^\text{FG}, D \right),
    \end{gathered}
\end{equation}
where $\text{diag}(\cdot)$ extracts the diagonal elements, and $\Sigma^\text{FG} \in \mathbb{R}^{L^\text{FG} \times D}$ is the diagonal matrix of singular values resulting from the SVD of the centered features $F^\text{FG} - \mu^{\text{FG}}$.
Intuitively, a large $\kappa^{\text{instance}} > 1$ indicates that the features are dominated by a few principal directions, whereas $\kappa^{\text{instance}} \approx 1$ suggests an approximately spherical distribution.
The same procedure is applied to $P$, with detailed derivations provided in the supplementary material.

As shown in \cref{tabs:2_misalignment} (c) and (d), the difference between $\mathcal{D}^\text{instance}$ and $\kappa^\text{instance}$ from $P$ and $F^\text{FG}$ is evident, indicating mismatches in feature scale and direction.
This discrepancy is not resolved with the projection layers preceding the cross-attention mechanisms; hence, the cross-attention struggles to capture the semantically meaningful relationships due to extreme misalignment.
Such large distributional gaps between prototypical tokens and support features not only hinder semantic correspondence during cross-attention, but also impede optimization \cite{santurkar2018does}.
This observation motivates us to project support features into a more coherent and structured representation space for effective attention-based prototype generation.

\section{Method}
\label{sec:methods}
\begin{figure*}[!t]
\centering
\includegraphics[width=1.\textwidth]{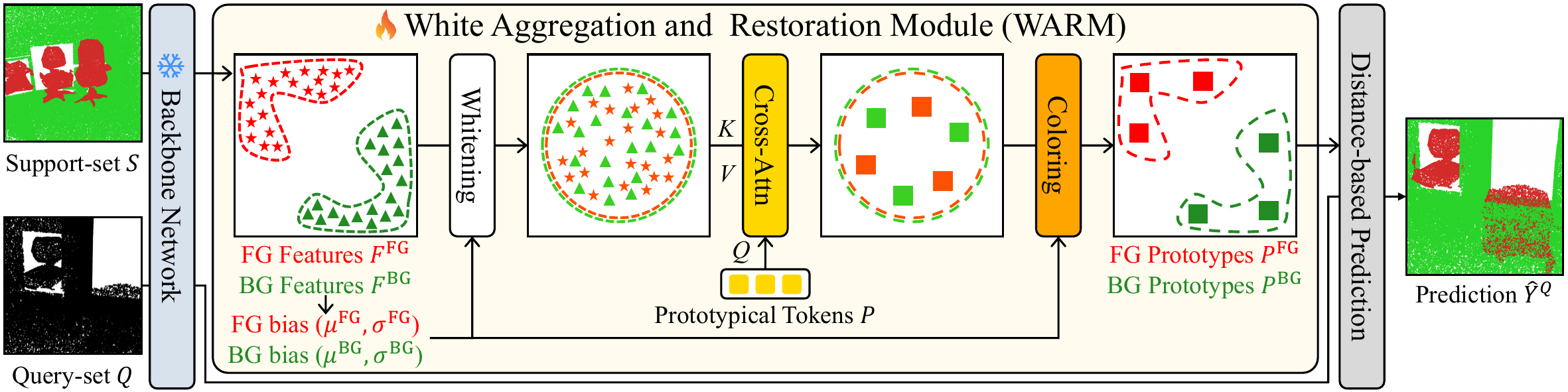}
\caption{
\textbf{Overall pipeline of WARM in the 1-way 1-shot setting.}
Support and query features are extracted using a pretrained network.
Within WARM, support features are divided into foreground and background components, which are whitened to align with the prototypical tokens and aggregated via cross-attention to generate prototypes.
The prototypes are then recolored to restore instance-specific statistics.
Finally, query points are classified based on their similarity to the prototypes.
}
\label{figs:2_main}
\end{figure*}

\subsection{Overview}
In this section, we present WARM, a cross-attention-based prototype generation framework for FS-PCS that incorporates a simple alignment strategy for improved semantic aggregation.
As shown in \cref{figs:2_main}, WARM comprises three stages: whitening, cross-attention, and coloring.
Support features are first whitened to decouple distributional biases and align them with prototypical tokens.
Cross-attention then aggregates the aligned features to form prototypes, which are subsequently colored with the instance-specific statistics of the features.
Finally, segmentation is performed by assigning each query point to its nearest prototype.
Overall, WARM achieves effective feature–prototype alignment while preserving essential instance characteristics.

\subsection{White Aggregation and Restoration Module}
\label{subsec:warm}
As detailed in \cref{subsec:attention_misalignment}, the vanilla cross-attention is insufficient to aggregate point features based on their semantic relationships.
To address this issue, we suggest the incorporation of alignment and restoration techniques with the cross-attention mechanism.
Specifically, for each class $C \in \{\text{FG}, \text{BG}\}$ in the support set, the covariance matrix $\mathcal{S}^C \in \mathbb{R}^{D \times D}$ is computed as:
\begin{equation}
    \label{eq:covariance}
    \mathcal{S}^C = \frac{1}{L^C-1} (F^C - \mu^C)^T (F^C - \mu^C).
\end{equation}
Then, we apply ZCA whitening \cite{bell1997independent, huang2018decorrelated} to $F^C$ as follows:
\begin{equation}
    \begin{split}
        \label{eq:whitening}
        Z^C = (F^C - \mu^C) \cdot (\mathcal{S}^C)^{-\frac{1}{2}},
    \end{split}
\end{equation}
where $Z^C \in \mathbb{R}^{L^C \times D}$ denotes the whitened features with zero mean and decorrelated channels, \textit{i.e.}, ${Z^C}^\top \cdot Z^C = I$.
By normalizing instance-specific statistics, the whitened features reside in a standardized space, where prototypical tokens are aligned with the point features.
With the whitened support features $Z^C$, we generate prototypes $\tilde{P}^C \in \mathbb{R}^{M \times D}$, which replace those in \cref{eq:vanilla_ca}, as follows:
\begin{equation}
    \tilde{P}^C_m = P_m + \text{CA}(P_m,Z^C).
    \label{whitened_ca}
\end{equation}
This facilitates stable aggregation based on semantics, as the attention mechanism no longer needs to account for instance-specific distributional variations that lead to isolated pairwise matching.

Although whitening discards statistics that would otherwise hinder alignment during attention, prototypes representing support features should retain these statistics to preserve the original expressiveness.
To this end, we introduce a coloring step after cross-attention.
This can be simply implemented as the inverse of the ZCA whitening in \cref{eq:whitening}, as follows:
\begin{equation}
    \begin{split}
        \label{eq:coloring}
        P^C = \tilde{P}^C \cdot (\mathcal{S}^C)^{\frac{1}{2}} + \mu^C,
    \end{split}
\end{equation}
where $P^C \in \mathbb{R}^{M \times D}$ denotes the final prototypes that incorporate complete information.
As such, we obtain representative prototypes that are derived by considering semantic relationships within point clouds while restoring instance-specific distributional statistics.

\subsection{Regularized Whitening for Stability}
\label{subsec:inversion_of_cov}
In order to apply whitening as in \cref{subsec:warm}, inverse square root of $\mathcal{S}^{C}$ is required, which is obtained by inverting the square root of its eigendecomposition:
\begin{equation}
\label{eq:cov_eigendecompose}
    (\mathcal{S}^{C})^{-\frac{1}{2}}
    = (V \Lambda V^T)^{-\frac{1}{2}} 
    = V \Lambda^{-\frac{1}{2}} V^T.
\end{equation}
The diagonal entries $\{\lambda_0, \lambda_1, ... \lambda_D\}$ of the diagonal matrix $\Lambda$ are the eigenvalues of $\mathcal{S}^C$ and the columns of $V$ are the corresponding eigenvectors.

However in FS-PCS, where the number of input points is dynamic, the covariance matrix $\mathcal{S}^C$ is possibly ill-conditioned in the case of sparse points due to linear dependencies.
In such cases, even a small variance in a certain direction that is most likely a noise, can be falsely amplified as a signal in the whitened space.
Therefore, to ensure stability and to prevent such noise-amplification, we regularize the eigenvalues:
\begin{equation}
\label{eq:eigen_reg}
    \Lambda \leftarrow (\mathbbm{1}_{\lambda > \nu}\lambda + \mathbbm{1}_{\lambda \leq \nu} \nu) \cdot I, \: \nu=\lambda_\text{median} \Bigl( 1+\sqrt{D/L^C} \Bigr)^2,
\end{equation}
where the threshold $\nu$ is set to the upper bound of estimated noise \cite{bai1993limit}, with $\lambda_\text{median}$ representing the median eigenvalue.
By thresholding the eigenvalues adaptively to the point cloud size, the inversion process is stabilized even in sparse scenes. 
Further details are provided in supplementary.

\subsection{Training Objective and Inference}
We adopt a basic segmentation approach, in which each query point is assigned the class of its nearest prototype.  
The prediction $\hat{Y}^Q_l$ for the $l$-th query point features $F^Q_l$ is determined by the distance to the nearest prototype within each class $C \in \{\text{FG}, \text{BG}\}$:
\begin{equation}
    \hat{Y}^Q_l = \underset{C}{\arg\min} \left( \min_{m} \lVert F^Q_l - P^C_{m} \rVert_2 \right).
    \label{eq:prediction}
\end{equation}
For supervision, we use a margin loss with the margin set to zero, to penalize only the wrong predictions:
\begin{equation}
    \mathcal{L}_{\mathrm{margin}} = \frac{1}{L} \sum_{l=1}^{L} 
    \Bigl( 
    \mathbbm{1}_{Y^Q_l =\text{FG}} 
    \max \left( d^\text{FG}_l - d^\text{BG}_l, 0 \right) 
    +
    \mathbbm{1}_{Y^Q_l =\text{BG}} 
    \max \left( d^\text{BG}_l - d^\text{FG}_l, 0 \right)
    \Bigr).
    \label{eq:margin_loss}
\end{equation}
where $\mathbbm{1}_\text{condition}$ is the indicator function, which equals 1 if the condition is true and 0 otherwise.
To further prevent the multi-prototypes from collapsing into trivial representations, we additionally apply a simplification loss \cite{lang2020samplenet}, as:
\begin{equation}
        \label{eq:simplication_loss}
        \!\!\!
        \mathcal{L}_{\mathrm{sim}} = 
        \frac{1}{N+1}
        \sum_{C}
        \Bigl(
        \frac{1}{L^C}\sum_{l=1}^{L^C} \underset{m}{\min}\:d^C_{lm}
        + \frac{1}{M}\sum_{m=1}^M \underset{l}{\min}\:d^C_{lm}
        + \underset{m}{\max}\:\underset{l}{\min}\:d^C_{lm}
        \Bigr),
\end{equation}
where $d^C_{lm}=\parallel F^C_{l} - P^C_{m} \parallel_2$.
The loss is computed separately for FG and BG.

As a result, the overall training objective is defined as:
\begin{equation}
    \mathcal{L} = \mathcal{L}_{\mathrm{margin}} + \alpha \cdot \mathcal{L}_{\mathrm{sim}},
\end{equation}
where $\alpha$ is a coefficient hyperparameter, set to 0.5.

\section{Experiments}
\label{sec:experiments}

\subsection{Experimental Results}

\subsubsection{Baselines.}
FS-PCS studies are conducted under two distinct strategies of preprocessing raw point clouds, 1) \textit{foreground oversampling} \cite{zhao2021few, ning2023boosting, PAPFZS3D} and 2) \textit{uniform sampling} \cite{an2024rethinking, an2024multimodality}.
As noted in \cite{an2024rethinking}, the former relies on ground-truth masks to oversample foreground regions, limiting its use in real-world settings where annotations are unavailable.
Therefore, we adopt and compare with the baselines under the uniform sampling protocol \cite{an2024rethinking, an2024multimodality}.
In addition, we report results for baselines originally developed under the foreground oversampling protocol \cite{zhao2021few, ning2023boosting, PAPFZS3D}, re-implemented within the uniform sampling setting, provided in \cite{an2024rethinking}.

\subsubsection{Datasets.}
We perform experiments on two benchmark datasets for FS-PCS: S3DIS \cite{armeni20163d} and ScanNet \cite{dai2017scannet}.
S3DIS comprises 271 indoor scenes with 12 semantic classes, and ScanNet contains 1,513 indoor scenes with annotations for 20 categories.
For both datasets, the classes are split into two folds, $S_0$ and $S_1$, for cross-validation, where they do not overlap with each other.
Following prior work \cite{zhao2021few}, each scene is divided into 1m$\times$1m blocks, with each block valid only if it contains $\geq5\%$ foreground points, with a 100 point minimum.
We adopt the same data preprocessing as in \cite{an2024rethinking}, where each block is voxelized into 0.02m grids, and uniformly sampled up to 20,480 points.

\subsubsection{Implementation Details.}
We use the encoder layers of Stratified Transformer \cite{lai2022stratified} as the feature extractor, with pretrained weights from \cite{an2024rethinking}.
For prototype generation, we use a single WARM layer with 100 prototypical tokens for each foreground and background, totaling 200 tokens.
In a multi-shot setting $K>1$, we generate prototypes for each sample and average them across shots as in \cite{snell2017prototypical}.
Evaluation is based on mean Intersection-over-Union (mIoU), and is performed on 1,000 episodes per class in the 1-way setting and 100 episodes per class combination for the 2-way setting to ensure stable results.

For training, AdamW optimizer with a learning rate of $1 \times 10^{-4}$ and a weight decay of 0.01 is applied.
Training is conducted for 10 and 20 epochs for the $N$-way 1-shot and 5-shot settings, respectively, where each epoch consists of 400 episodes.
The learning rate is decayed by a factor of 0.1 at 60\% and 80\% of the total number of epochs. 
All experiments are performed on an RTX A6000 GPU.

\begingroup
\setlength{\tabcolsep}{1.35pt}
\renewcommand{\arraystretch}{1.0}
\begin{table*}[t]
\caption{
\textbf{Quantitative result on S3DIS in mIoU~(\%).}
Combinations of 1/2-way, 1/5-shot setting are reported, for splits $S_0$ and $S_1$, including their mean. 
The best and second-best are \textbf{bolded} and \underline{underlined}, respectively.
$\dagger$ denotes reproduced performance without foreground leakage with Stratified Transformer~\cite{lai2022stratified} as backbone, provided in~\cite{an2024rethinking}.
$*$ denotes methods that utilize additional modality. 
}
\vspace{-0.2cm}
\centering
\fontsize{6.5}{8}\selectfont
\begin{tabular}{l|ccc|ccc|ccc|ccc| c}
    \toprule
    \multirow{2}{*}{Method} & \multicolumn{3}{c|}{1-way 1-shot} & \multicolumn{3}{c|}{1-way 5-shot} & \multicolumn{3}{c|}{2-way 1-shot} & \multicolumn{3}{c|}{2-way 5-shot} & \multirow{2}{*}{Overall} \\ 
    \cline{2-13}
    & $S_0$ & $S_1$ & Mean & $S_0$ & $S_1$ & Mean & $S_0$ & $S_1$ & Mean & $S_0$ & $S_1$ & Mean & \\
    \midrule
    AttMPTI$^\dagger$~\cite{zhao2021few} & 36.32 & 38.36 & 37.34 & 46.71 & 42.70 & 44.71 & 31.09 & 29.62 & 30.36 & 39.53 & 32.62 & 36.08 & 37.12 \\ 
    QGE$^\dagger$~\cite{ning2023boosting}& 41.69 & 39.09 & 40.39 & 50.59 & 46.41 & 48.50 & 33.45 & 30.95 & 32.20 & 40.53 & 36.13 & 38.33 & 39.86 \\
    QGPA$^\dagger$~\cite{PAPFZS3D} & 35.50 & 35.83 & 35.67 & 38.07 & 39.70 & 38.89 & 25.52 & 26.26 & 25.89 & 30.22 & 32.41 & 31.32 & 32.94 \\
    COSeg~\cite{an2024rethinking} & 46.31 & 48.10 & 47.21 & 51.40 & 48.68 & 50.04 & 37.44 & 36.45 & 36.95 & 42.27 & 38.45 & 40.36 & 43.64 \\
    \rowcolor{gray!20} MM-FSS$^*$~\cite{an2024multimodality} & 49.84 & \textbf{54.33} & \underline{52.09} & 51.95 & 56.46 & 54.21 & 41.98 & \textbf{46.61} & 44.30 & 46.02 & \textbf{54.29} & 50.16 & 50.19 \\
    FPS + \textit{min-dist.} & \underline{51.35} & 45.68 & 48.52 & \underline{68.43} & \underline{61.51} & \underline{64.97} & \underline{44.48} & 33.00 & \underline{38.74} & \underline{58.22} & 45.52 & \underline{51.87} & \underline{51.03} \\
    \midrule
    \rowcolor{maroon!10} WARM (ours) & 
    \textbf{60.16} & \underline{50.50} & \textbf{55.33} & \textbf{72.23} & \textbf{62.66} & \textbf{67.45} & 
    \textbf{50.91} & \underline{38.34} & \textbf{44.63} & \textbf{61.46} & \underline{48.95} & \textbf{55.21} & \textbf{55.66} \\
    \bottomrule
\end{tabular}
\label{tabs:3_s3dis}

\medskip
\setlength{\tabcolsep}{1.35pt}
\renewcommand{\arraystretch}{1.0}
\caption{
\textbf{Quantitative result on ScanNet in mIoU~(\%).}
Experimental setting is the same as~\cref{tabs:3_s3dis}, performed on ScanNet~\cite{dai2017scannet}. }
\vspace{-0.2cm}
\centering
\fontsize{6.5}{8}\selectfont
\begin{tabular}{l|ccc|ccc|ccc|ccc|c}
    \toprule
    \multirow{2}{*}{Method} & \multicolumn{3}{c|}{1-way 1-shot} & \multicolumn{3}{c|}{1-way 5-shot} & \multicolumn{3}{c|}{2-way 1-shot} & \multicolumn{3}{c|}{2-way 5-shot} & \multirow{2}{*}{Overall} \\
    \cline{2-13}
    & $S_0$ & $S_1$ & Mean & $S_0$ & $S_1$ & Mean & $S_0$ & $S_1$ & Mean & $S_0$ & $S_1$ & Mean \\
    \midrule
    AttMPTI$^\dagger$~\cite{zhao2021few} & 34.03 & 30.97 & 32.50 & 39.09 & 37.15 & 38.12 & 25.99 & 23.88 & 24.94 & 30.41 & 27.35 & 28.88 & 31.11 \\ 
    QGE$^\dagger$~\cite{ning2023boosting} & 37.38 & 33.02 & 35.20 & 45.08 & 41.89 & 43.49 & 26.85 & 25.17 & 26.01 & 28.35 & 31.49 & 29.92 & 33.66 \\
    QGPA$^\dagger$~\cite{PAPFZS3D} & 34.57 & 33.37 & 33.97 & 41.22 & 38.65 & 39.94 & 21.86 & 21.47 & 21.67 & 30.67 & 27.69 & 29.18 & 31.19 \\
    COSeg~\cite{an2024rethinking} & \underline{41.73} & \underline{41.82} & \underline{41.78} & 48.31 & 44.11 & 46.21 & 28.72 & 28.83 & 28.78 & 35.97 & 33.39 & 34.68 & 37.86\\ 
    \rowcolor{gray!20} MM-FSS$^*$~\cite{an2024multimodality} & \textbf{46.08} & \textbf{43.37} & \textbf{44.73} & \textbf{54.66} & 45.48 & 50.07 & \textbf{43.99} & \textbf{34.43} & \textbf{39.21} & \textbf{48.86} & 39.32 & \textbf{44.09} & \textbf{44.53} \\
    FPS + \textit{min-dist.} & 37.95 & 33.66 & 35.81 & 52.77 & \underline{47.72} & \underline{50.25} & 30.13 & 27.06 & 28.60 & 43.41 & \underline{39.35} & 41.38 & 39.01 \\ 
    \midrule
    \rowcolor{maroon!10} WARM (ours) & 
    41.16 & 39.10 & 40.13 & \underline{53.40} & \textbf{49.04} & \textbf{51.22} & \underline{30.27} & \underline{30.02} & \underline{30.15} & \underline{45.02} & \textbf{41.61} & \underline{43.32} & \underline{41.21} \\
    \bottomrule
\end{tabular}
\label{tabs:4_scannet}
\end{table*}
\vspace{-0.3cm}
\endgroup

\subsubsection{Quantitative Result.}
We compare WARM with existing baselines under 1/2-way and 1/5-shot settings.
As shown in \cref{tabs:3_s3dis}, WARM achieves state-of-the-art performance for uni-modal methods on the S3DIS dataset.
This superiority is mostly maintained on the ScanNet dataset in \cref{tabs:4_scannet} for uni-modal models, although it has lower performance in most scenarios compared to multi-modal method such as MM-FSS \cite{an2024multimodality}.
MM-FSS exploits text and image features in addition to point clouds for more comprehensive pretraining \cite{li2022language}, whereas the main contribution of WARM is focused on the prototype construction, rather than feature quality.

Given the simplicity of WARM compared to heavy decoder-based models \cite{an2024multimodality, an2024rethinking}, its competitive performance highlights the importance of careful prototype design.
Moreover, the strong results achieved by `FPS + \textit{min-dist.}' further question the necessity of complex decoders.
These findings suggest that future work should place greater emphasis on developing more expressive and robust prototype representations, as explored in other few-shot downstream tasks \cite{zhang2022feature, lee2025temporal}.

\begin{figure}[!t]
    \begin{minipage}[T]{0.56\columnwidth}
        \centering
        \includegraphics[width=0.95\textwidth]{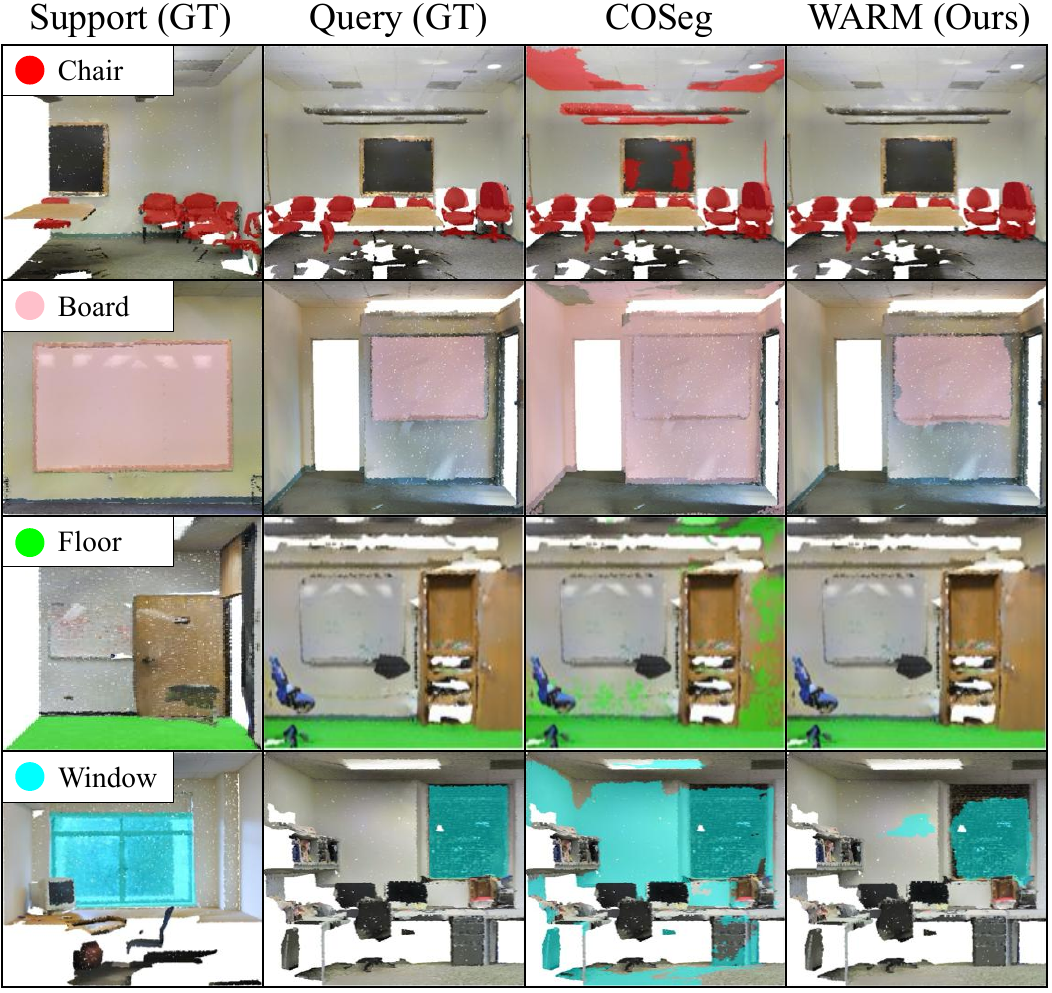}
        \caption{
        Qualitative results of COSeg~\cite{an2024rethinking} and WARM on 1-way 1-shot setting of the S3DIS~\cite{armeni20163d} dataset.
        Each row denotes a single test episode, where ground-truths~(GT) and predictions are highlighted with their corresponding class color.
        }
        \label{figs:3_qualitative}
    \end{minipage}
    \hfill
    \begin{minipage}[T]{0.43\columnwidth}
        \vspace{-0.4cm}
        \setlength{\tabcolsep}{2.55pt}
\renewcommand{\arraystretch}{1.1}
\captionof{table}{
Component ablation study.
\textbf{N}, \textbf{W}, and \textbf{R} denote normalization, whitening, and coloring.
Normalization means scaling the features to zero mean and unit variance along each channel dimension.
Dist$(Q, K)$ indicates the distance between queries and keys as in \cref{eq:q_k_distance}.
}
\smallskip
\centering
\scriptsize
\begin{tabular}{c|ccc|c|c}
    \toprule
    & \textbf{N} & \textbf{W} & \textbf{R}& Dist$(Q,K)$ $\downarrow$ & mIoU~(\%) \\
    \midrule
    (a) & & & & 288.81 & 47.29 \\ % CA
    (b) & \checkmark & & & 13.30 & 14.18 \\ % STD_CA_NC
    (c) & & \checkmark & & 13.14 & 10.83 \\ % White_CA_NC
    (d) & \checkmark & & \checkmark & 13.72 & 57.59 \\ % STD_CA
    (e) & & \checkmark & \checkmark & \textbf{13.13} & \textbf{60.16} \\ 
    \bottomrule
\end{tabular}
\label{tabs:5_component}

        \vspace{-0.1cm}
        \setlength{\tabcolsep}{4.0pt}
\renewcommand{\arraystretch}{1.1}
\captionof{table}{
Loss ablation study.
}
\scriptsize
\centering
\begin{tabular}{l|cc|c}
\toprule  
 & $\mathcal{L}_\text{margin}$ & $\mathcal{L}_\text{sim}$ & mIoU~(\%) \\
\midrule
Na\"ive CA  & \checkmark & \checkmark & 47.29 \\
\midrule
\multirow{3}{*}{WARM}  & \checkmark &  & 59.34 \\
  &  & \checkmark & 58.74 \\
  & \checkmark & \checkmark & \textbf{60.16} \\
\bottomrule
\end{tabular}
\label{tabs:6_loss}
    \end{minipage}
    \vspace{-0.3cm}
\end{figure}

\subsubsection{Qualitative Result.}
In addition to the quantitative results, we display qualitative results in \cref{figs:3_qualitative}.
Compared to COSeg \cite{an2024rethinking}, WARM consistently reduces mispredictions and increases correct classifications across episodes, while maintaining a simpler segmentation architecture.
Along with quantitative results, this demonstrates the importance of prototype integrity, rather than focusing on complex decoders.

\subsection{Ablation Study}
Experiments are conducted under the 1-way 1-shot setting with $S_0$ of S3DIS \cite{armeni20163d}.

\subsubsection{Component Ablation.}
\label{subsubsec:component_ablation}
\cref{tabs:5_component} presents a component-wise ablation study verifying the effectiveness of each alignment step.
In (a), the na\"ive cross-attention (CA) suffers from severe misalignment between query and key features, as reflected in the large Dist$(Q, K)$.
We address this via whitening-based alignment, with comparisons against normalization, an alternative alignment method.
As shown in (b), normalization standardizes features to unit variance per channel, substantially reducing misalignment, but remains sub-optimal as it preserves the covariance structure.
In contrast, whitening (c) enforces an isotropic distribution by removing the covariance, further decreasing the misalignment.
However, while both methods enable cross-attention to capture semantic relationships, they also strip essential distributional characteristics, causing prototypes to lose the original instance's inherent semantics, as reflected in their low mIoU.
Therefore, coloring (d–e) should be accompanied to restore these properties after alignment.
The full model in (e) with whitening and coloring achieves the best performance, confirming that both feature alignment and semantic restoration are necessary for faithful prototype representation.

\subsubsection{Loss Ablation.} 
As shown in \cref{tabs:6_loss}, WARM substantially outperforms CA, even with only the task objective $\mathcal{L}_{\text{margin}}$, indicating that WARM's architecture itself mitigates attention degradation.
Combining $\mathcal{L}_{\text{margin}}$ and $\mathcal{L}_{\text{sim}}$ achieves the best result, suggesting complementary effects.

\begin{figure}[!t]
    \begin{minipage}[T]{0.38\columnwidth}
        \vspace{-0.5cm}
        \setlength{\tabcolsep}{2.5pt}
\renewcommand{\arraystretch}{1.1}
\captionof{table}{
Quantitative comparison of attention maps.
Entropy (normalized to the range $[0,1]$) measures the uniformity of each attention distribution.
Diversity is computed as the inverse of average similarity among the attention maps of the prototypical tokens, reflecting how distinctly they focus on different regions.
}
\centering
\scriptsize
\begin{tabular}{l|c|c}
    \toprule
    Method & Entropy & Diversity \\
    \midrule
    Cross-Attention & 0.2079 & 0.8824 \\
    WARM & 0.8387 &  0.6442 \\
    \bottomrule
\end{tabular}
\label{tabs:7_diversity_entropy}
    \end{minipage}
    \hfill
    \begin{minipage}[T]{0.6\columnwidth}
        \centering
        \includegraphics[width=\textwidth]{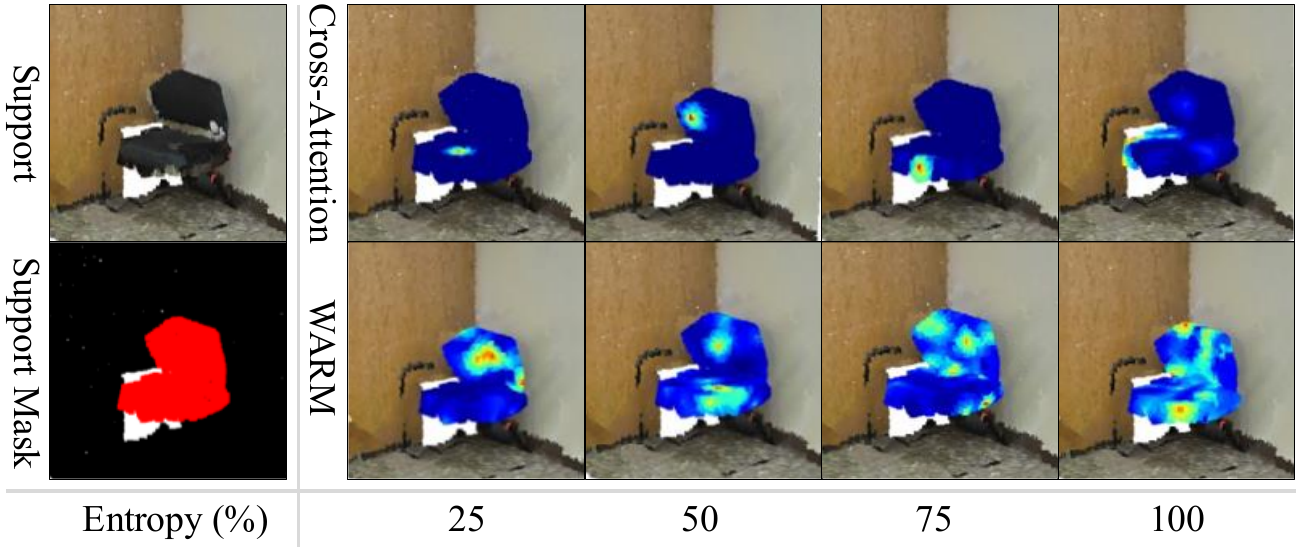}
        \captionof{figure}{
        Qualitative comparison of attention maps. 
        To ensure a fair assessment, each attention map is sampled based on its respective ranking within the entropy score distribution. 
        }
        \label{figs:4_attention_map}
    \end{minipage}
    \vspace{-0.3cm}
\end{figure}

\subsubsection{Semantic Attention by Whitening.}
We validate that whitening enables prototypical tokens to aggregate features by semantic relationship, rather than spatial proximity in the projected space, through quantitative and qualitative analysis.
We define two metrics: entropy, quantifying attention distribution uniformity (normalized to $[0,1]$), and diversity, quantifying inter-prototype distinctiveness as the inverse of average attention-map similarity.
As shown in \cref{tabs:7_diversity_entropy}, na\"ive cross-attention exhibits low entropy and high diversity, indicating that prototypes attend to features close in projection space but lacking semantic relevance, causing attention to collapse into point-level aggregation without semantic grouping.
WARM, by contrast, achieves substantially higher entropy alongside a moderate but sufficient diversity score, indicating that each prototype attends broadly and consistently to semantically related regions, yielding richer and more coherent representations.
Qualitative results in \cref{figs:4_attention_map} corroborate this: na\"ive attention is spatially narrow, whereas WARM captures semantically coherent point groupings.
Visualizations are sampled by entropy percentile rather than cherry-picked results.

\subsubsection{Compatibility with FS-PCS Methods.}
To evaluate the versatility of WARM as a general-purpose prototype generation module, we integrate it into existing FS-PCS frameworks in the uniform sampling protocols, such as COSeg \cite{an2024rethinking} and MM-FSS \cite{an2024multimodality}.
Specifically, we replace their standard FPS-based prototype generation with WARM.
Notably, the mIoU of COSeg increases from 46.31\% to 59.26\%, representing a substantial 12.95\% improvement. 
Similarly, MM-FSS shows a steady gain, rising from 49.84\% to 52.62\%.
These results demonstrate that WARM is a robust, plug-and-play solution that effectively substitutes conventional heuristic-based sampling.
Additional experiments on the foreground oversampling baselines are provided in the supplementary.

\begin{figure}[!t]
    \begin{minipage}[T]{0.52\columnwidth}
        \centering
\captionof{table}{
Parametric capacity ablation.
% While marginal improvements occur with capacity expansion as shown in (a), it falls behind attention-based (d).
% On the other hand, structural improvement of (d) exhibits substantial improvements, when compared to (c) with with the same capacity.
% Other methods with larger parameters has lower performance, as shown in (b).
% Results (a-c) shows that increasing the parameter has minimal effect in performance, while high performance in (d) indicates that structural design is more important.  
Results (a-c) indicate that increasing the parameter has minimal effect on performance, whereas the substantial gain in (d) suggests that structural design is the primary determinant of performance.
}
% \medskip
\label{tabs:8_capacity}
\centering
\scriptsize
\begin{tabular}{l|l|c|c|c}
\toprule
& \multirow{2}{*}{Method} & \multirow{2}{*}{mIoU} & Backbone & Extra 
\\
&  &  & param. \# & param. \# 
\\
\midrule
\multirow{2}{*}{(a)} & \multirow{2}{*}{FPS $\!$+$\!$ \textit{min-dist.}} & 51.35 & 3.91M & - 
\\
 & & 52.34 & 5.16M & - 
\\
\midrule 
\multirow{2}{*}{(b)} & COSeg \cite{an2024rethinking} & 46.31 & 3.91M & 2.20M 
\\
& MM-FSS \cite{an2024multimodality} & 49.84 & 7.55M & 7.22M
\\
\midrule 
(c) & CA & 47.29 & 3.91M & 1.13M
\\
\midrule
\rowcolor{maroon!10}(d) & WARM & \textbf{60.16} & 3.91M & 1.13M
\\
\bottomrule
\end{tabular}
    \end{minipage}
    \hfill
    \begin{minipage}[T]{0.45\columnwidth}
        \includegraphics[width=\columnwidth]{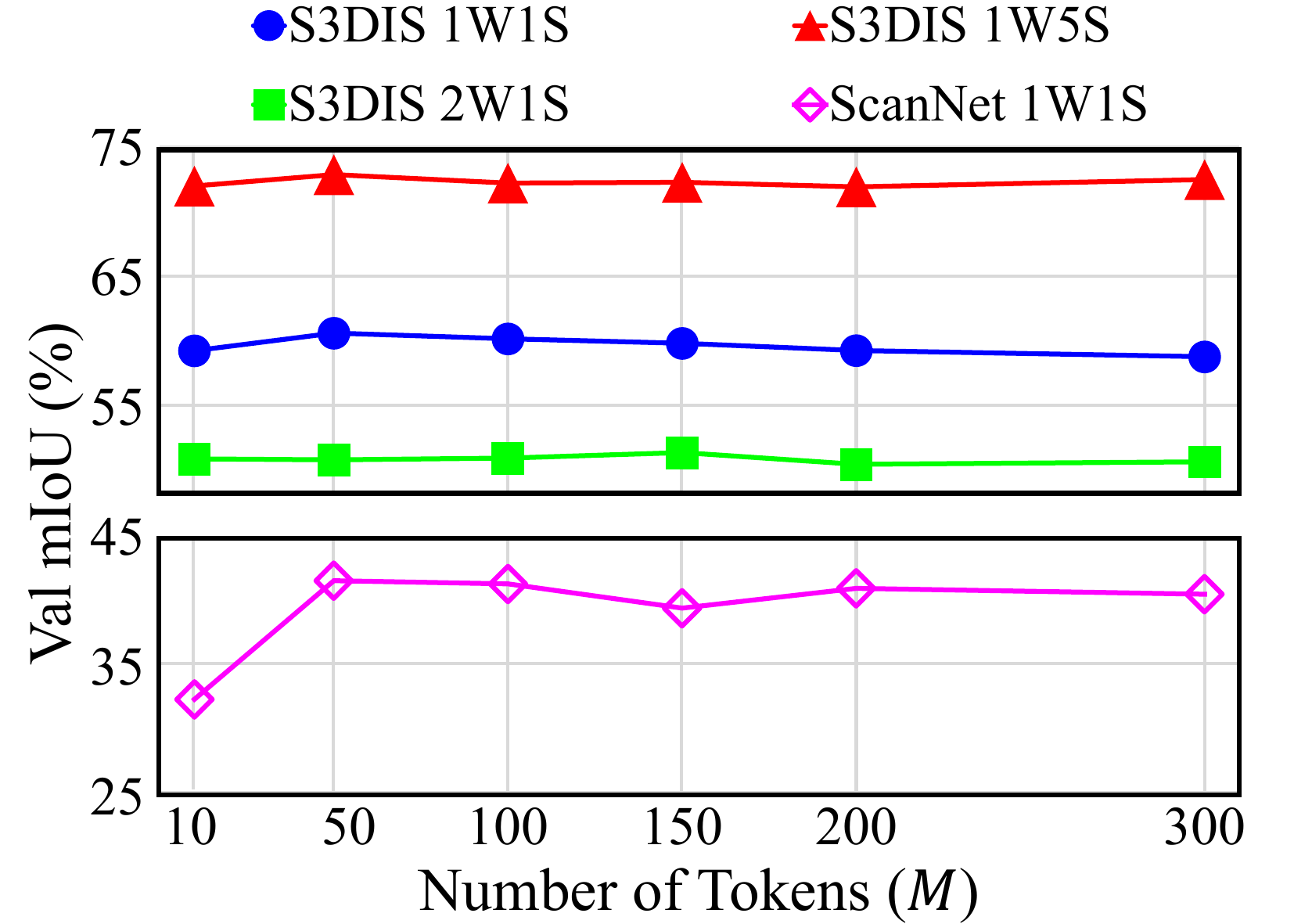}
        \caption{
        Number of prototypical tokens ($M$) ablation for different settings.
        % For S3DIS dataset, the performance remains stable across various number of tokens.
        % However, dataset with larger number of classes such as ScanNet, exhibits drops in performance for extremely small $M$.
        }
        \label{figs:5_tokens_M}
    \end{minipage}
\vspace{-0.2cm}
\end{figure}

\subsubsection{Parametric Capacity Ablation.}
To verify that WARM's gain comes from its structured design rather than additional parametric capacity, we analyze the performance of existing methods with respect to their parameter counts.
In \cref{tabs:8_capacity}, (a) increasing the backbone capacity of `FPS+\textit{min-dist}.' from 3.91M to 5.16M brings only marginal gains in performance, while (b) existing FPS-based methods underperform WARM despite using comparable or larger backbones with much larger additional modules.
These results imply that additional module capacity does not directly improve performance.
Conclusively, (c) shows that even the same structure as WARM performs poorly without whitening/coloring, while (d) confirms superior performance.
To sum up, WARM's gain mainly comes from resolving the misalignment between learnable tokens and support features through whitening and coloring, rather than from simply adding parameters.

\subsubsection{Number of Tokens.}
We provide ablation on $M$ in \cref{figs:5_tokens_M} across different datasets and settings.
For S3DIS, the performance remains stable when varying $M$, indicating that increasing the number of tokens does not consistently improve performance.
However in ScanNet, the performance largely drops at small value of $M\! = \! 10$, likely due to its larger number of FG classes requiring higher representational capacity, and stabilizing once $M\! \ge\! 50$.
In conclusion, the results suggest that more complex datasets may require a sufficient number of tokens for adequate representational capacity, although performance becomes stable once $M$ exceeds a moderate threshold.

\subsubsection{Memory and Runtime Overhead.}
We compare peak memory and runtime of WARM against FPS-based prototype generation.
For constructing 100 prototypes, we vary the point count ($L=$1K$\sim$25K; $D{=}100$) and the feature dimension ($D=$100$\sim$2K; $L{=}1$K), one factor at a time.
As shown in \cref{tabs:9_overhead}, WARM uses slightly more peak memory due to the attention parameters, but is significantly faster than FPS.
This is because FPS is inherently sequential, whereas\setlength{\intextsep}{2pt}      % vertical space above/below wrap
\setlength{\columnsep}{12pt}
\setlength{\tabcolsep}{1.8pt}
\begin{wraptable}{r}{0.6\columnwidth}
\caption{Runtime and memory overhead of WARM compared to FPS.
}
\label{tabs:9_overhead}
\centering
\scriptsize
\begin{tabular}{l|c|ccc|ccc}
\toprule
\multirow{2}{*}{Overhead} & \multirow{2}{*}{Method} & \multicolumn{3}{c|}{Point Count}  & \multicolumn{3}{c}{Feature Dimension}
\\
& & 1K & 5K & 25K & 100 & 500 & 2K
\\
\midrule
Memory $\downarrow$ & FPS & 39.4 & 196.5 & 983.6 & 39.4 & 193.8 & 772.9
\\
(MB) & WARM & 42.6 & 204.3 & 1016.1 & 42.6 & 216.5 & 974.2
\\
\midrule
Runtime $\downarrow$ & FPS & 9.9 & 20.8 & 78.3 & 10.1 & 20.5 & 87.8
\\
(ms) & WARM & 2.8 & 2.8 & 6.0 & 2.8 & 16.2 & 85.6
\\
\bottomrule
\end{tabular}
\end{wraptable}
\noindent WARM is dominated by GPU-parallel batched matrix multiplications.
Overall, the computational efficiency of the attention module outweighs the additional cost incurred by whitening, thereby maintaining the practical runtime feasibility of WARM.

% \input{figs/6_optimization_group}
% \subsection{Optimization Benefits of Whitening}
% Beyond its effect in aligning queries and keys in the cross-attention, whitening provides additional benefits in optimization.
% As illustrated in \cref{figs:6_optimization}, WARM achieves faster convergence by maintaining higher gradient magnitudes during training.
% This results from effective gradient descent achieved by decorrelated channels \cite{ahmad2024correlations}.
% Despite the increased gradient flow, the training remains stable, suggesting that whitening enhances both optimization speed and robustness.

\section{Conclusion}
In this paper, we investigated prototype generation for FS-PCS, addressing the limitations of conventional approaches in FS-PCS settings.
We observed that even a widely adopted mechanism struggles to bridge the distributional gap when trained with limited samples, motivating our design of an enhanced cross-attention module that incorporates whitening and coloring transformations.
Through this design, we obtained representative prototypes by decoupling detrimental factors while preserving originality, thus achieving competitive performance across FS-PCS benchmarks even without the aid large decoders.
Pointing out the underexplored impact of prototype generation, we hope our work offers a complementary direction for FS-PCS.

\section*{Acknowledgements}
This work was supported in part by MSIT/IITP (No. RS-2022-II220680, RS-2020-II201821, RS-2019-II190421, RS-2024-00459618, RS-2024-00360227, RS-2024-00437633, RS-2024-00437102, RS-2025-25442569), MSIT/NRF (No. RS-2024-00357729), KNPA/KIPoT (No. RS-2025-25393280), and SEMES-SKKU collaboration funded by SEMES.

\bibliographystyle{splncs04}
\bibliography{main}

% ---------------------------------------------------------------
% Supplementary
\clearpage

\title{Supplementary Material of \\White Aggregation and Restoration for \\Few-shot 3D Point Cloud Semantic Segmentation}
\titlerunning{White Aggregation and Restoration for FS-PCS}
\author{Jiyun Im \and
SuBeen Lee \and
Miso Lee \and
Jae-Pil Heo*}

\def\thefootnote{*}\footnotetext{Corresponding author}

\authorrunning{J. Im et al.}

\institute{Sungkyunkwan University \\
\email{\{bbangsil0110, leesb7426, dlalth557, jaepilheo\}@skku.edu}}

\maketitle

\setcounter{section}{0}
\renewcommand{\thesection}{\Alph{section}}
\setcounter{figure}{0}
\renewcommand{\thefigure}{A\arabic{figure}}
\setcounter{table}{0}
\renewcommand{\thetable}{A\arabic{table}}
\setcounter{equation}{16}

\section{Derivation of Instance-wise Anisotropy}

We elaborate the derivation of Eq.~(6) from Sec.~4.2.
To begin with, the singular value decomposition (SVD) of the mean centered support foreground (FG) features is denoted as:
\begin{equation}
    \label{eq:svd}
    F^\text{FG} - \mu^\text{FG} = U \Sigma^\text{FG} V^T,
\end{equation}
where the rows of $U$ and the columns of $V$ each represents left and right singular vectors, and $\Sigma^\text{FG} \in \mathbb{R}^{L^\text{FG} \times D}$ representing a rectangular diagonal matrix with singular values on the diagonal.
Specifically, the singular values $\left[\Sigma^\text{FG}_{11}, \Sigma^\text{FG}_{22}, \cdots,\Sigma^\text{FG}_{rr}\right]$ are in descending order, \textit{i.e.} $\Sigma^\text{FG}_{11} \ge \Sigma^\text{FG}_{22} \ge \cdots \ge \Sigma^\text{FG}_{rr}$, with $r=\min(L^\text{FG},D)$.

\section{Numerical Stability and Robustness of Whitening}
The superscript $C$ is omitted for brevity in the following section.

\subsection{Details on Regularized Whitening}
\label{subsec:inversion_of_cov_sup}
We provide additional details on the regularization method described in Sec.~5.3.
While thresholding in Eq.~(12) is effective in cases where $L > D$, in other cases where $L \le D$, the inversion itself becomes undefined due to zero eigenvalues.
In such cases, we compute the inverse only on the submatrix corresponding to the positive eigenvalues and set the remaining diagonal entries of the inverse to zero as follows:

\begin{equation}
\label{eq:eigen_reg_inv}
    \Lambda^{-\frac{1}{2}} \leftarrow \mathbbm{1}_{\lambda > 0} \cdot (\lambda + \epsilon)^{-\frac{1}{2}} \cdot I,
\end{equation}
where $\epsilon$ is a small positive value.

Additionally, when $L \le D$, the threshold $\nu=\lambda_\text{median} \cdot  ( 1 + \sqrt{D/L} )^2 $ in Eq.~(12) becomes unsuitable as the factor $D / L$ grows rapidly as $L$ decreases, leading to excessive regularization.
For continuous regularization without reciprocal factors, we adopt the threshold $\nu$ corresponding to $L = D$ for $L \le D$, which is formally written as:
\begin{equation}
\label{eq:eigen_reg_2}
    \begin{gathered}
    \Lambda \leftarrow (\mathbbm{1}_{\lambda > \nu}\lambda + \mathbbm{1}_{0 < \lambda \leq \nu} \nu) \cdot I,
    \\
    \nu=\min(\lambda_\text{max}, \: \lambda_\text{median} \cdot ( 1 + \sqrt{D/D} )^2)
    = \min(\lambda_\text{max}, \: \lambda_\text{median} \cdot 4),
    \end{gathered}
\end{equation}
where $\lambda_\text{max}$ denotes the largest eigenvalue.
The corresponding inverse is derived as shown in \cref{eq:eigen_reg_inv} using the updated $\Lambda$. 

%%%%%%%%%%%%%%%%%%%%%%%%%%%%

\setlength{\columnsep}{12pt}
\setlength{\tabcolsep}{4pt}
\begin{table}[!t]
\caption{Ablation study on threshold $\nu$.
Experiments are conducted on the test episodes of $S_0$ of S3DIS \cite{armeni20163d}, partitioned by the number of points.
The entry $\epsilon$ denotes clipping eigenvalues to a small positive constant $(\approx 2.22 \times 10^{-16})$, while $\lambda_\text{mean}$ is the mean of positive eigenvalues.
The columns $\kappa$ and reg. quantifies the directional imbalance of channels and the ratio of regularized eigenvalues below the threshold $\nu$, averaged across episodes.
}
\centering
\label{tabs:A1_regularization}
\begin{tabular}{c|c|ccc}
    \toprule
     \# of points & threshold $\nu$ & mIoU (\%) & $\kappa$ & reg. (\%)
     \\
     \midrule
     \multirow{4}{*}{$ L > D$} & $\epsilon$ ($\approx2.22 \times 10^{-16}$) & 60.37 & 3903.57 & -
     \\
     & $\lambda_\text{mean}$ & 57.46 & 8.48 & 91.73
     \\
     & $\lambda_\text{median}$ & 60.87 & 73.74 & 50.00
     \\
     & \cellcolor{maroon!10}$\lambda_\text{median}\cdot(1 + \sqrt{D/L})^2$ & \cellcolor{maroon!10}\textbf{61.07} & \cellcolor{maroon!10}39.63 & \cellcolor{maroon!10}66.39
     \\
     \midrule
     \multirow{5}{*}{$ L \le D$} & $\epsilon$ ($\approx2.22 \times 10^{-16}$) & 58.65 & 4.54$\times 10^9$ & -
     \\
     & $\lambda_\text{mean}$ & 55.62 & 6.80 & 90.31
     \\
     & $\lambda_\text{median}$ & 59.75 & 51.31 & 50.38
     \\
     & $\lambda_\text{median}\cdot(1 + \sqrt{D/L})^2$ & 59.01 & 22.82 & 75.60
     \\
     & \cellcolor{maroon!10}$\lambda_\text{median} \cdot 4$ & \cellcolor{maroon!10}\textbf{59.79} & \cellcolor{maroon!10}25.63 & \cellcolor{maroon!10}71.03
     \\
    \bottomrule
\end{tabular}
\end{table}

\subsection{Ablation Study on Regularized Whitening}

Dynamic eigenvalue regularization is essential in the whitening process, since the condition of the covariance matrix varies according to the number of support points.
In particular, while a large threshold effectively suppresses noise directions, it risks discarding potentially useful information.
Conversely, a threshold that is too small fails to filter out noise.
Therefore, selecting an optimal threshold is critical to balancing noise reduction and information retention.

To evaluate our regularization method described in Sec.~5.3 and \cref{subsec:inversion_of_cov_sup}, we compare it against alternative thresholding strategies.
Specifically, we use three criteria: mIoU (\%), covariance conditioning (quantified by $\kappa$ in Eq.~(6)), and the ratio of regularized eigenvalues, which reflects the aggressiveness of the threshold.
As shown in \cref{tabs:A1_regularization}, thresholding relative to the eigenvalues results in a significantly lower $\kappa$ compared to $\epsilon$ regularization.
This indicates that thresholding better conditions the covariance matrix, leading to robust whitening with reduced noise amplification.
However, a lower $\kappa$ only reflects improved conditioning of the covariance matrix and does not account for potential information loss due to aggressive regularization.
Although $\lambda_\text{mean}$ yields the best conditioning, it clips more than 90\% of the eigenvalues, indicating a higher risk of information loss.
The lower performance exhibited by $\lambda_\text{mean}$ supports this observation, suggesting a trade-off between robustness and discriminative power.
This insight motivates our choice of $\nu$ in the $L \le D$ case.
While the threshold inspired by \cite{bai1993limit} is effective for larger $L$, it can become overly conservative when $L$ is small.
The improvement in performance along with fewer clipped eigenvalues, provides empirical support for this reasoning.
When comparing $\lambda_\text{median}$ with our method, the latter yields consistently smaller $\kappa$, indicating improved robustness.
Overall, our method achieves a balance between robustness to noise and preservation of information.

\subsection{Implementation Details for Numerical Stability}
We additionally provide the implementation of WARM for reference.
Although we consistently expressed our method with eigenvalue decomposition for better understanding, computing the covariance matrix is generally avoided in practice due to potential precision loss.
Instead, applying SVD directly to the centered features avoids the loss of precision, and produce identical results.
The Eq.~(7) can be refactored using \cref{eq:svd} as follows:
\begin{equation}
    \begin{split}
        \label{eq:cov_svd}
        \mathcal{S}
        &= \frac{1}{L-1} (F - \mu)^T (F - \mu) \\
        &= \frac{1}{L-1} \left( U \Sigma V^T \right)^T \left( U \Sigma V^T \right) \\
        % &= \frac{1}{L-1} V (\Sigma)^T(\Sigma) V^T
        &= \frac{1}{L-1} V \Sigma^T\Sigma V^T
        = V \Lambda V^T.
    \end{split}
\end{equation}
Concretely, $\Lambda$ is denoted as:
\begin{equation}
    \Lambda=\frac{1}{L-1} \Sigma^T\Sigma.
\end{equation}

\section{Additional Experiments}

\begin{figure}[!t]
    \begin{minipage}[T]{0.57\columnwidth}
        \captionof{table}{
        Extended comparison of vanilla cross-attention (CA), normalization/de-normalization (ND) and whitening/coloring (WARM).
        }
        % \medskip
        \centering
        \renewcommand{\arraystretch}{1.0}
        \setlength{\tabcolsep}{3.3pt}
        \begin{tabular}{l|c|c|c|c}
        \toprule
        Method & mIoU & $\text{Dist}(P, F)$ & $\mathcal{D}^\text{inst.}$ & $\kappa^\text{inst.}$ \\
        \midrule
        % $P$ & - & - & 1.12 & 10.41 \\
        % \midrule
        CA & 47.29 & 288.81 & 95.40 & $1.2\times10^4$ \\ 
        ND & 57.59 & 13.72 & 13.50 & $2.9\times10^5$ \\ 
        WARM & \textbf{60.16} & \textbf{13.13} & \textbf{7.31} & $\mathbf{2.8\times10^3}$ \\
        \bottomrule
        \end{tabular}
        \label{tabs:A2_normalization}
    \end{minipage}
    \hfill
    \begin{minipage}[T]{0.39\columnwidth}
    \captionof{table}{
    Extended compatibility ablation with foreground oversampling method, VIP-Seg \cite{wang2025reasoning}.
    }
    \medskip
        \setlength{\tabcolsep}{5pt} % Default value: 6pt
        \renewcommand{\arraystretch}{1.1} % Default value: 1
        \centering
        \begin{tabular}{l|c|c}
        \toprule
        Method & 1W1S & 1W5S \\
        \midrule
        VIP-Seg \cite{wang2025reasoning} & 76.99 & 77.05 \\
        + WARM & \textbf{78.51} & \textbf{80.55} \\
        \bottomrule
        \end{tabular}
        \label{tabs:A3_vipseg}
    \end{minipage}
\end{figure}

\subsection{Comparison with Normalization/De-normalization (ND)}
We further analyze ND using metrics in Tab.~2 of Sec.~4.2, on $S_0$ of S3DIS dataset.
In \cref{tabs:A2_normalization}, while ND reduces the query-key distance $\text{Dist}(P, F)$ and instance dispersion $\mathcal{D}$, it exhibits severe anisotropy $\kappa$. 
Such anisotropy means that the learnable tokens concentrate the support features along a few principal directions, rather than distributing them across the full channel space, which limits the representational capacity.
In contrast, WARM mitigates $\kappa$ by decorrelating channels, improving representational capacity and mIoU.

\subsection{Compatibility with Foreground Oversampling Methods}
To strictly evaluate the practicality of WARM as a plug-and-play module, we also attach WARM to recent method in the foreground oversampling method such as VIP-Seg \cite{wang2025reasoning}.
\cref{tabs:A3_vipseg} reports the average mIoU on S3DIS over $S_0$ and $S_1$ for 1-way 1/5-shot settings.
The results verify that WARM is also complementary to foreground oversampling approaches and effective as a plug-and-play module.

\subsection{Optimization Benefits of Whitening}
Beyond its effect in aligning queries and keys in the cross-attention, whitening\begin{wrapfigure}{r}{0.6\columnwidth}
    % \vspace{-0.3cm}
    \includegraphics[width=0.6\columnwidth]{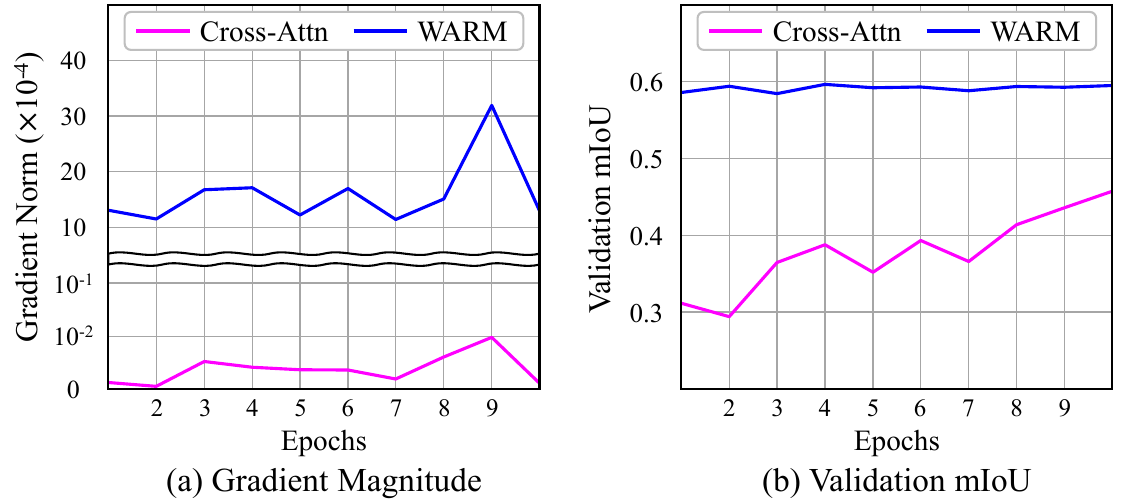}
    \caption{
    Training stability and acceleration comparison of WARM and na\"ive cross-attention.
    (a) shows the gradient magnitudes, while (b) demonstrates saturation in validation mIoU throughout training.
    }
    \label{figs:A1_optimization}
    % \vspace{-0.3cm}
\end{wrapfigure}\noindent provides additional benefits in optimization.
As illustrated in \cref{figs:A1_optimization}, WARM achieves faster convergence by maintaining higher gradient magnitudes during training.
This results from effective gradient descent achi-
eved by decorrelated channels \cite{ahmad2024correlations}.
Despite the increased gradient flow, the training remains stable, suggesting that whitening enhances both optimization speed and robustness.

\end{document}